\documentclass[runningheads]{llncs}

\usepackage[final,year=2024,ID=4474]{eccv}

\usepackage{eccvabbrv}

\usepackage{graphicx}
\usepackage{booktabs}
\usepackage{algorithm}
\usepackage{algorithmic}

\usepackage{multirow}

\usepackage[accsupp]{axessibility}  

\usepackage[pagebackref,breaklinks,colorlinks,citecolor=eccvblue]{hyperref}

\usepackage{orcidlink}

\usepackage{fancyhdr}
\makeatletter
\let\ps@plain\ps@fancy 
\makeatother

\begin{document}

\title{Promoting SAM for Camouflaged Object Detection \\via Selective Key Point-based Guidance} 

\titlerunning{Abbreviated paper title}

\author{Guoying Liang  ,Su Yang}

 \institute{}

\maketitle

\begin{abstract}
  Big model has emerged as a new research paradigm that can be applied to various down-stream tasks with only minor effort for domain adaption. Correspondingly, this study tackles Camouflaged Object Detection (COD) leveraging the Segment Anything Model (SAM). The previous studies declared that SAM is not workable for COD but this study reveals that SAM works if promoted properly, for which we devise a new framework to render point promotions: First, we develop the Promotion Point Targeting Network (PPT-net) to leverage multi-scale features in predicting the probabilities of camouflaged objects' presences at given candidate points over the image. Then, we develop a key point selection (KPS) algorithm to deploy both positive and negative point promotions contrastively to SAM to guide the segmentation. It is the first work to facilitate big model for COD and achieves plausible results experimentally over the existing methods on 3 data sets under 6 metrics. This study demonstrates an off-the-shelf methodology for COD by leveraging SAM, which gains advantage over designing professional models from scratch, not only in performance, but also in turning the problem to a less challenging task, that is, seeking informative but not exactly precise promotions.
\end{abstract}
\begin{figure*}[t]
 \centering
 \begin{subfigure}{0.32\linewidth}
  \includegraphics[scale=0.169]{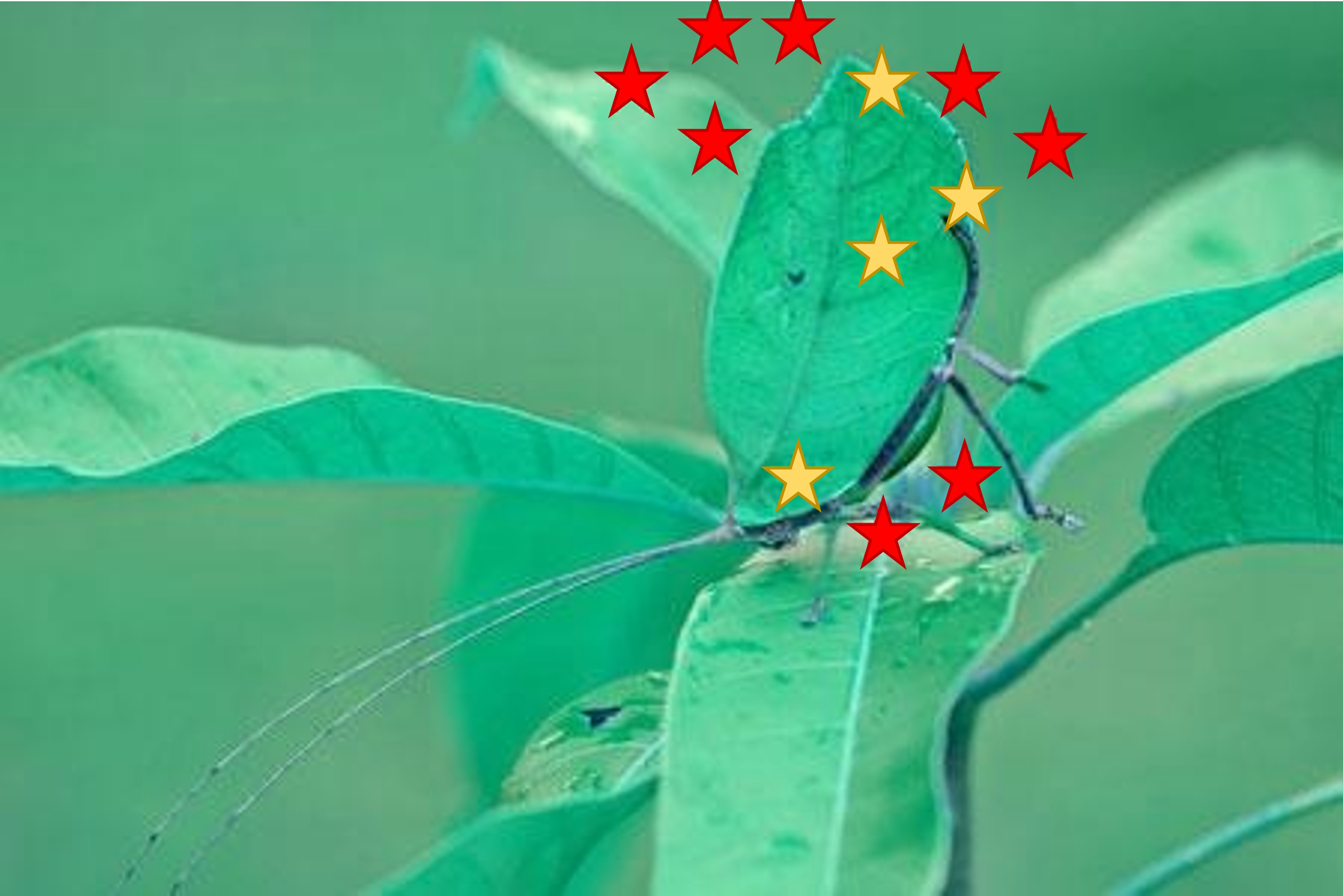}
  \label{fig:short-a}
 \end{subfigure}
 \vspace{-0.145in}
 \hspace{0.001in}
 \begin{subfigure}{0.32\linewidth}
  \includegraphics[scale=0.169]{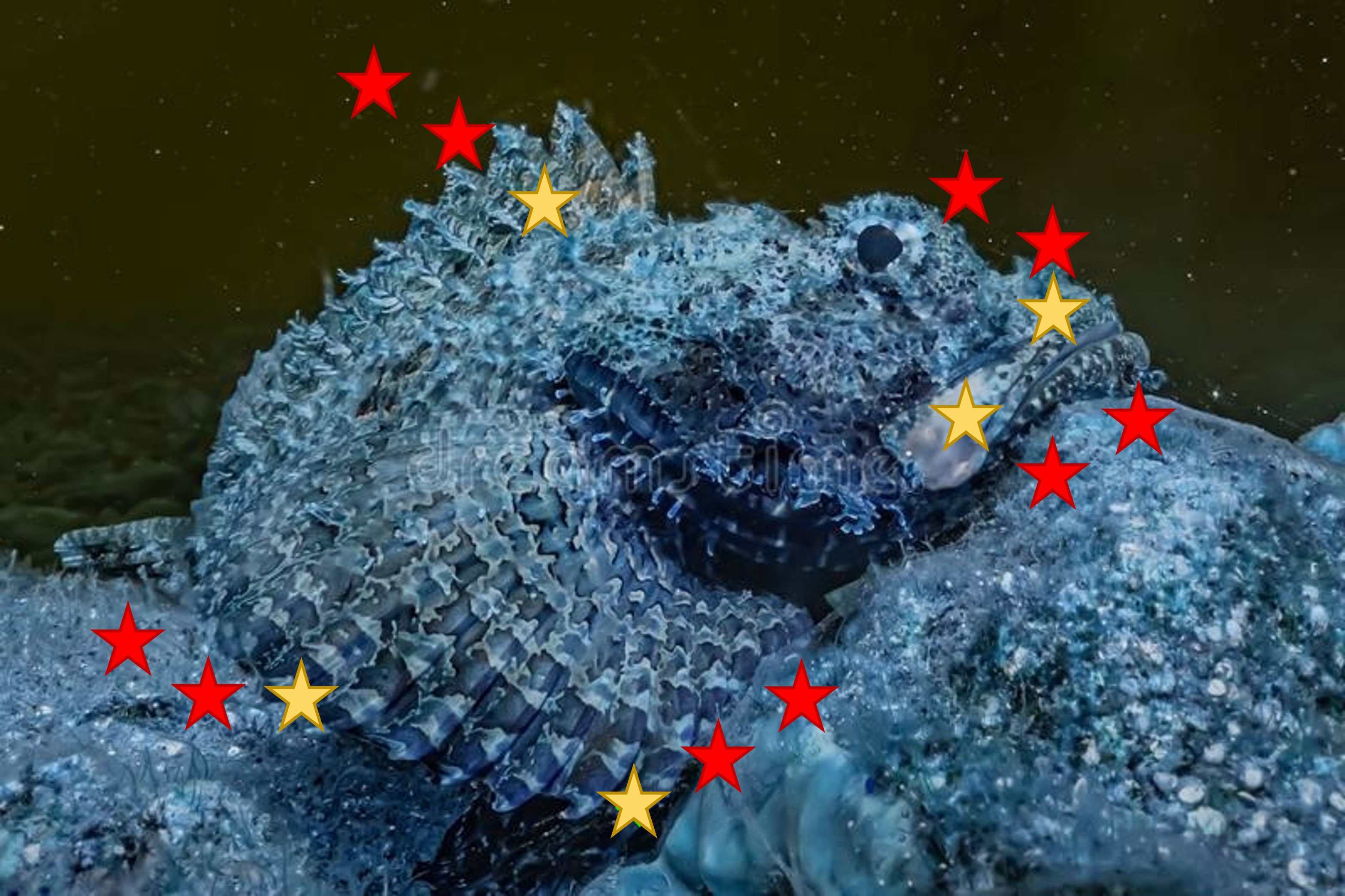}
  \label{fig:short-a}
 \end{subfigure}
 \hspace{0.001in}
 \begin{subfigure}{0.32\linewidth}
  \includegraphics[scale=0.169]{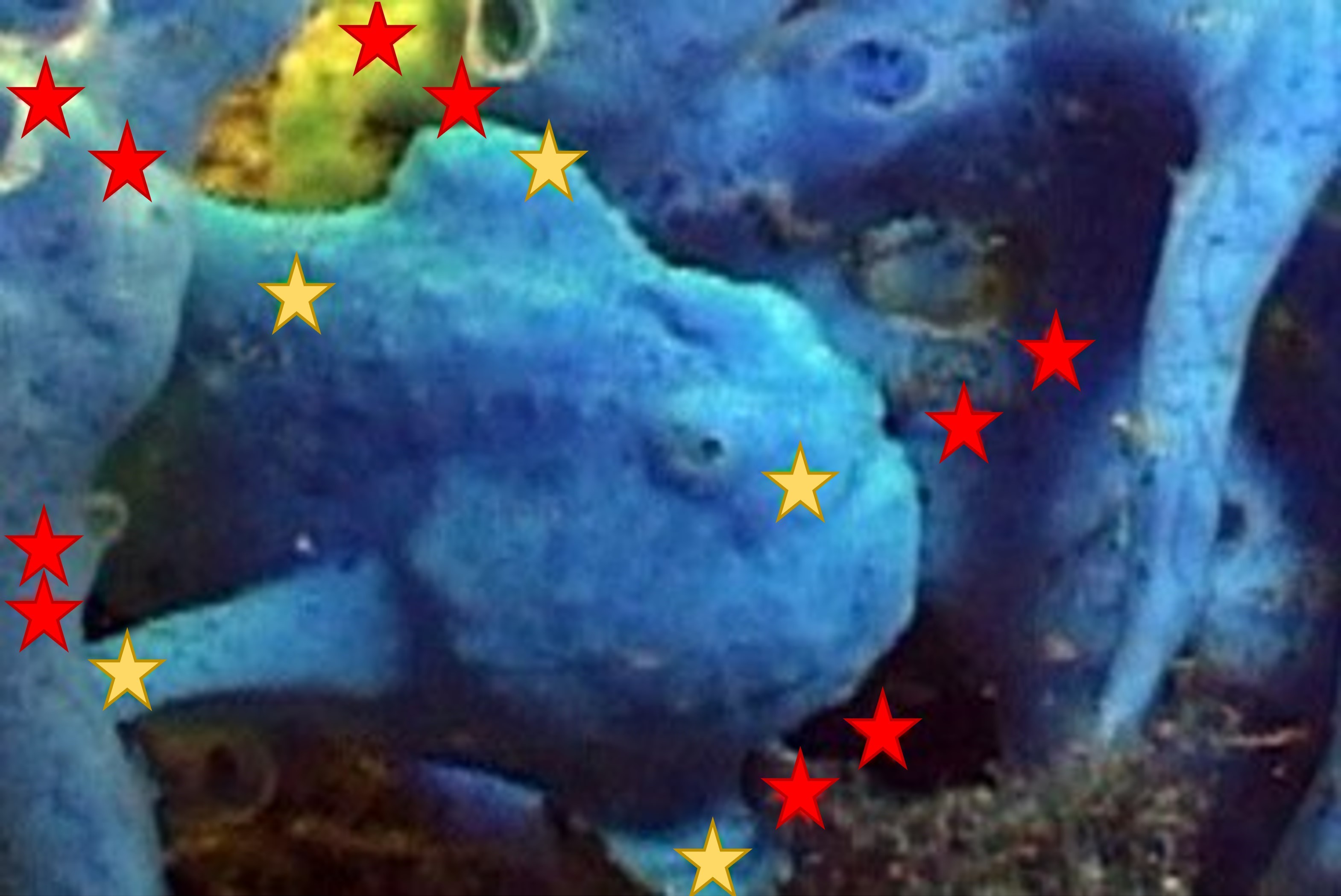}
  \label{fig:short-a}
 \end{subfigure}
 \hspace{0.001in}
 \begin{subfigure}{0.32\linewidth}
  \includegraphics[scale=0.169]{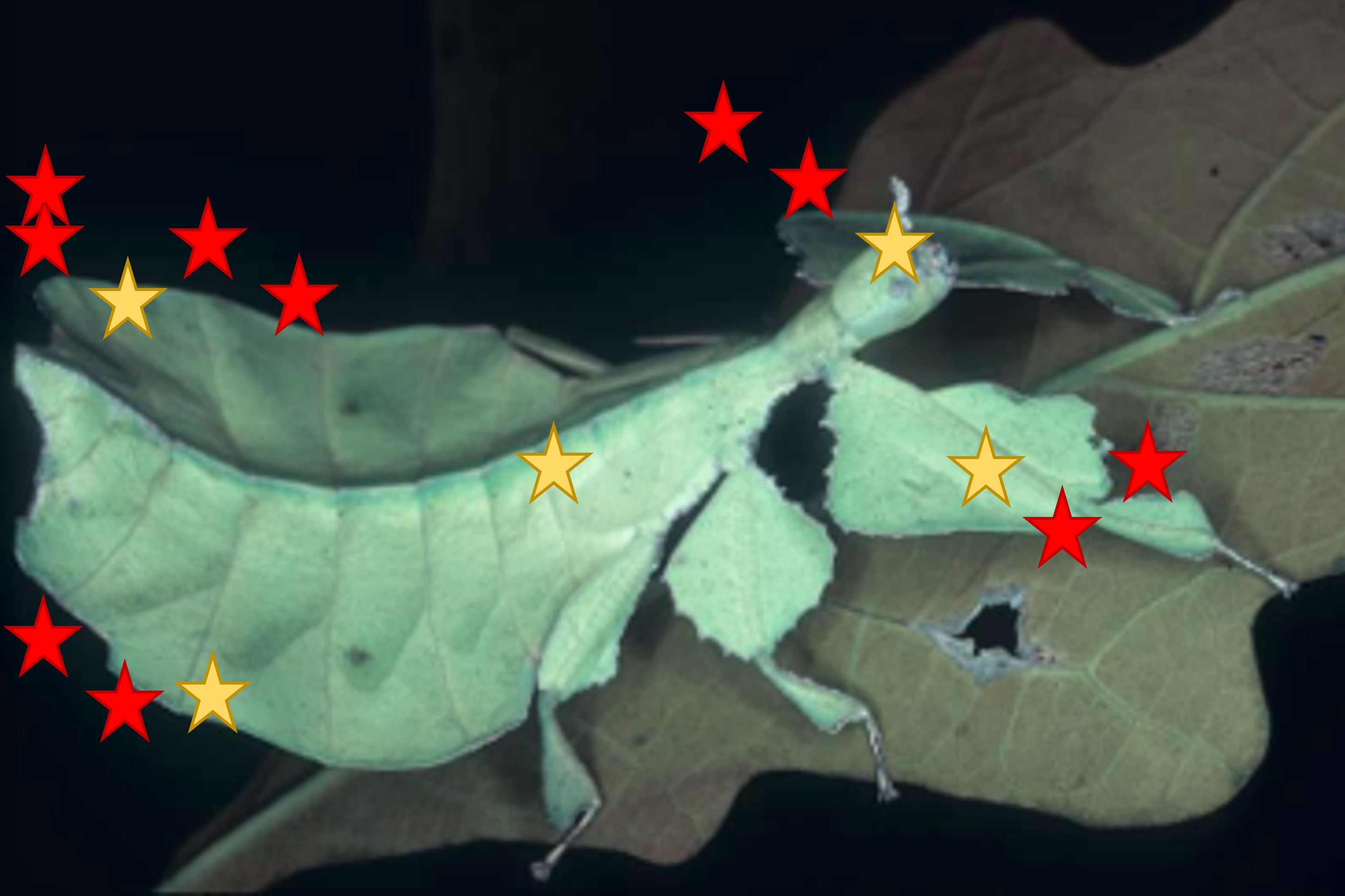}
  \label{fig:short-a}
 \end{subfigure}
 \hspace{0.001in}
 \begin{subfigure}{0.32\linewidth}
  \includegraphics[scale=0.169]{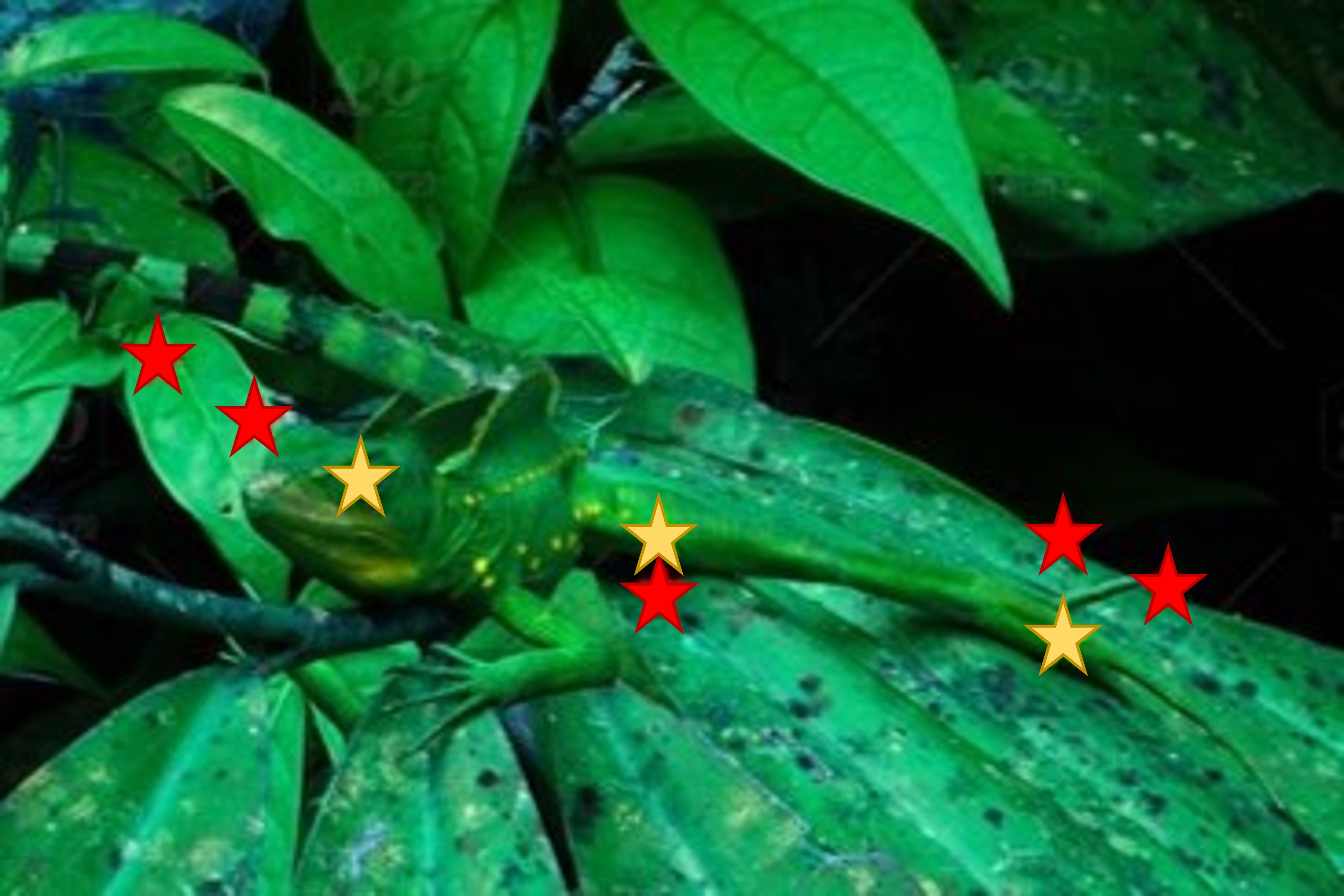}
  \label{fig:short-a}
 \end{subfigure}
 \hspace{0.001in}
 \vspace{-0.145in}
 \begin{subfigure}{0.32\linewidth}
  \includegraphics[scale=0.169]{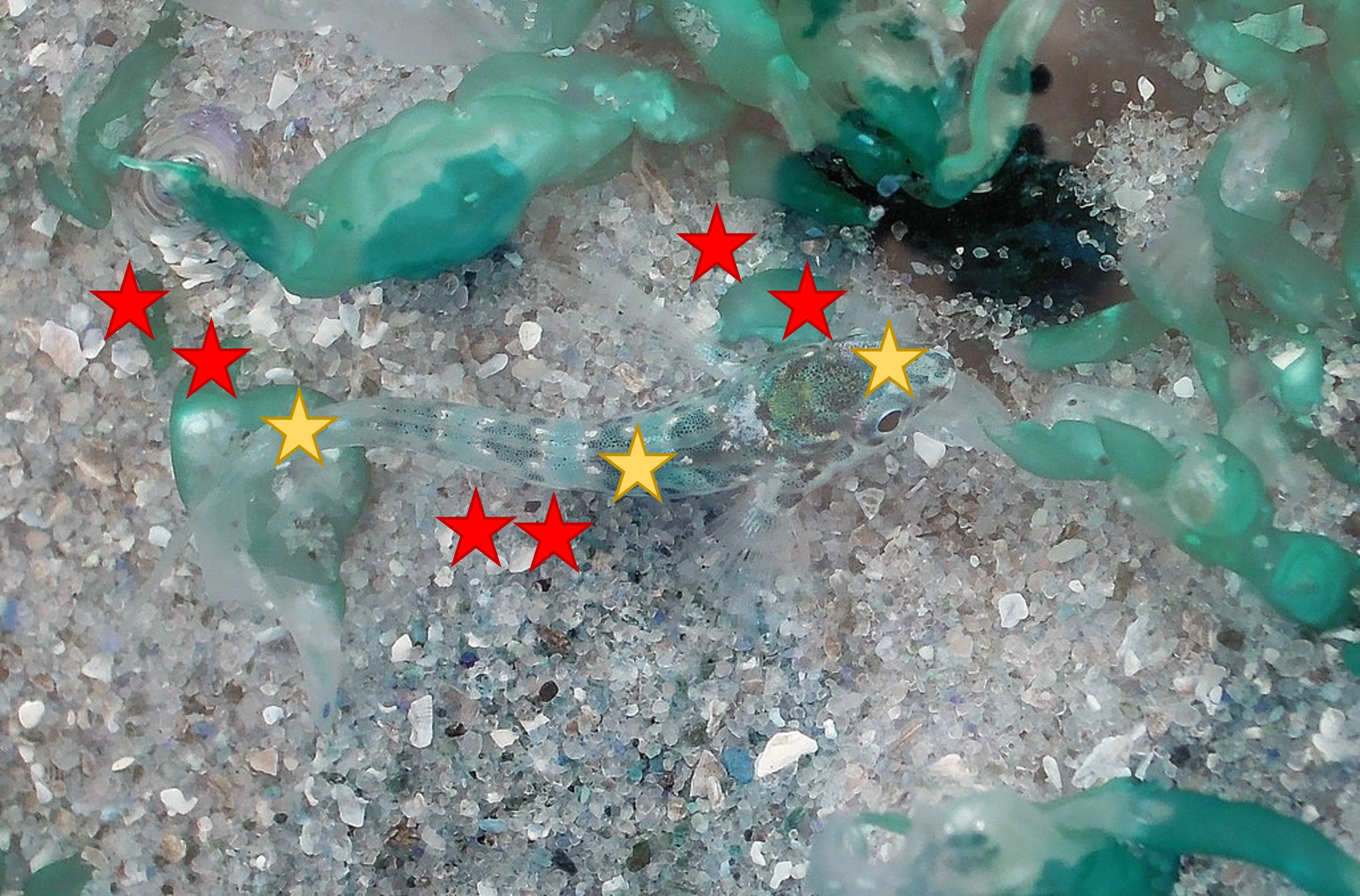}
  \label{fig:short-a}
 \end{subfigure}
 \begin{subfigure}{0.32\linewidth}
  \includegraphics[scale=0.169]{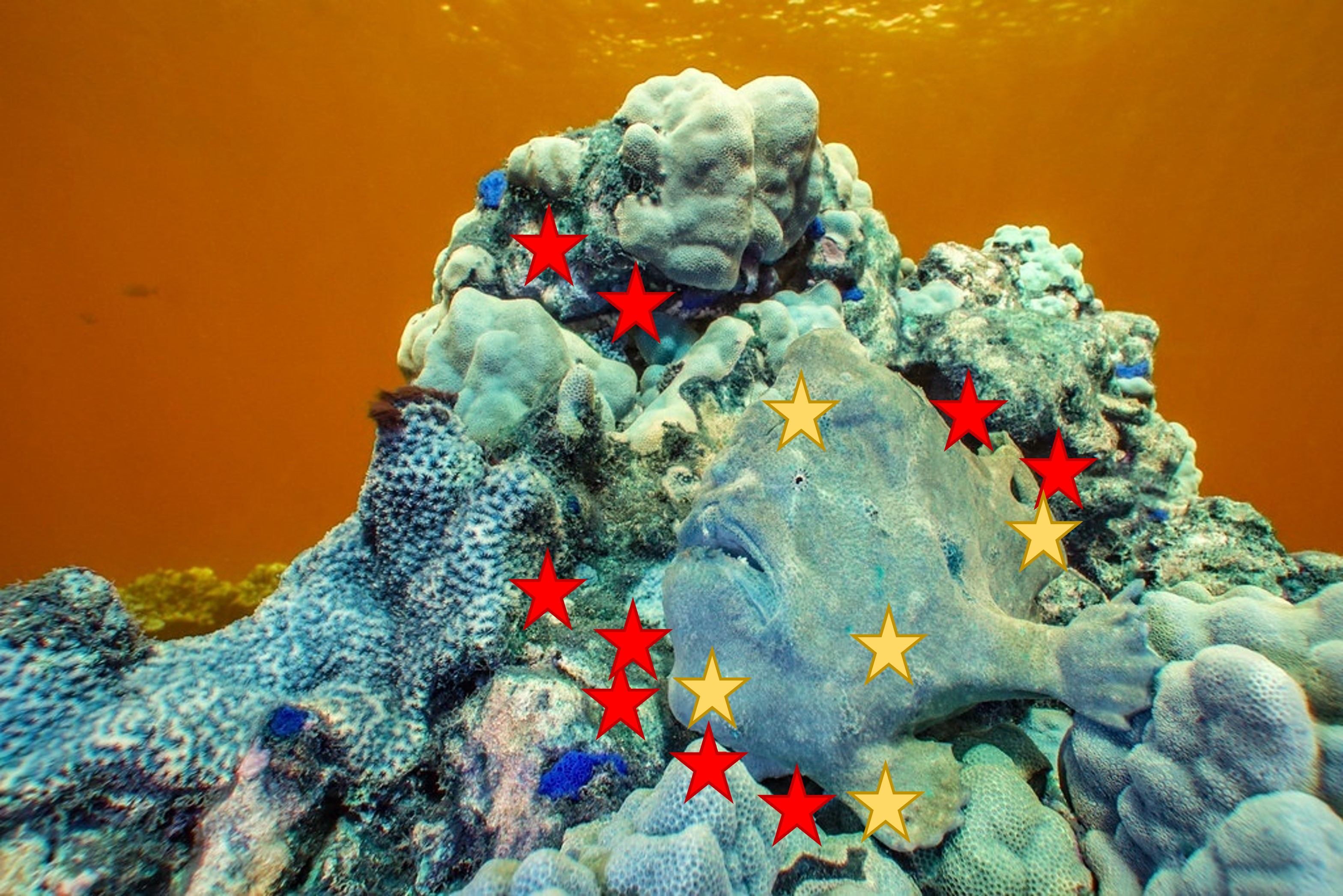}
  \label{fig:short-a}
 \end{subfigure}
 \hspace{0.001in}
 \begin{subfigure}{0.32\linewidth}
  \includegraphics[scale=0.169]{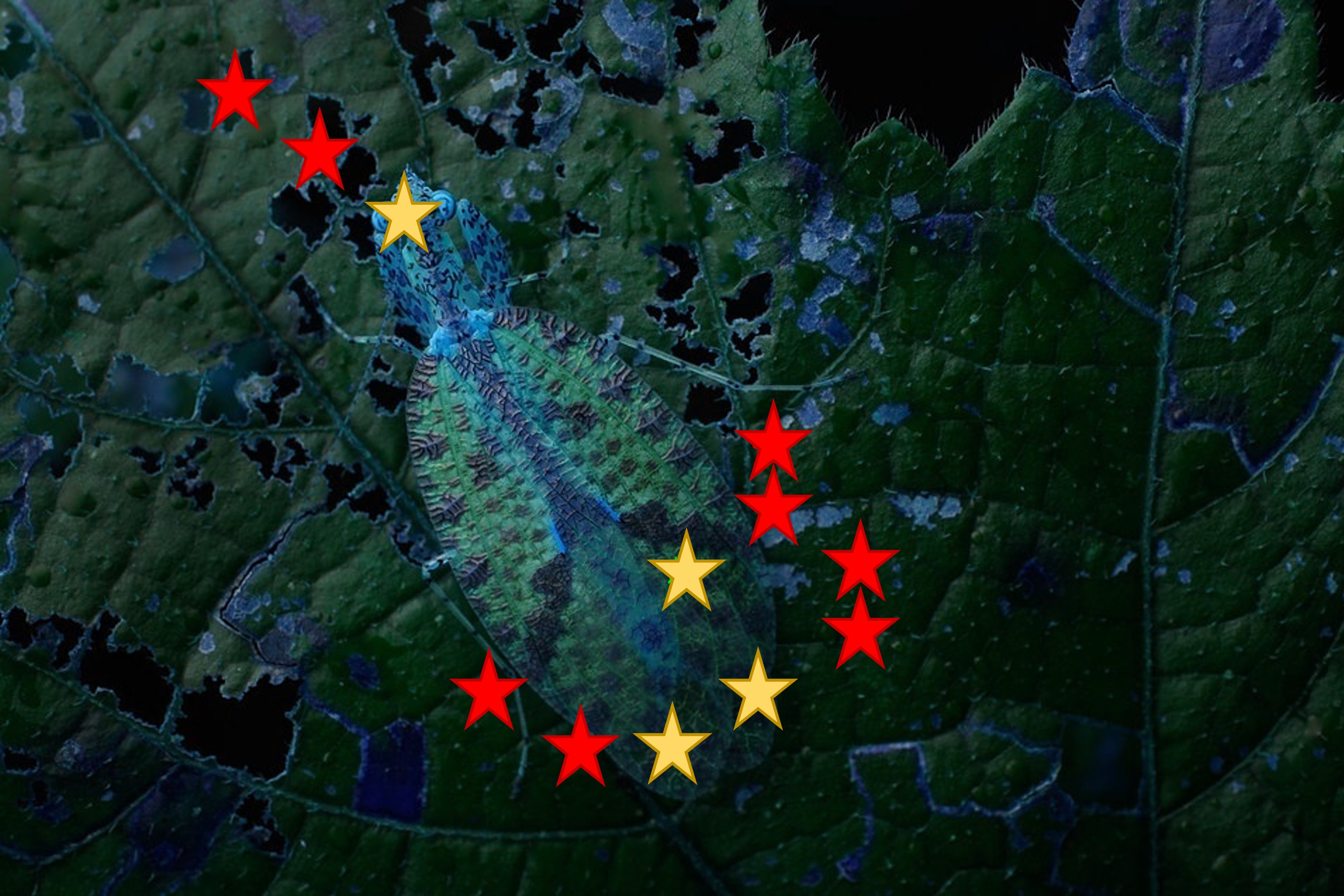}
  \label{fig:short-a}
 \end{subfigure}
 \hspace{0.001in}
 \begin{subfigure}{0.32\linewidth}
  \includegraphics[scale=0.169]{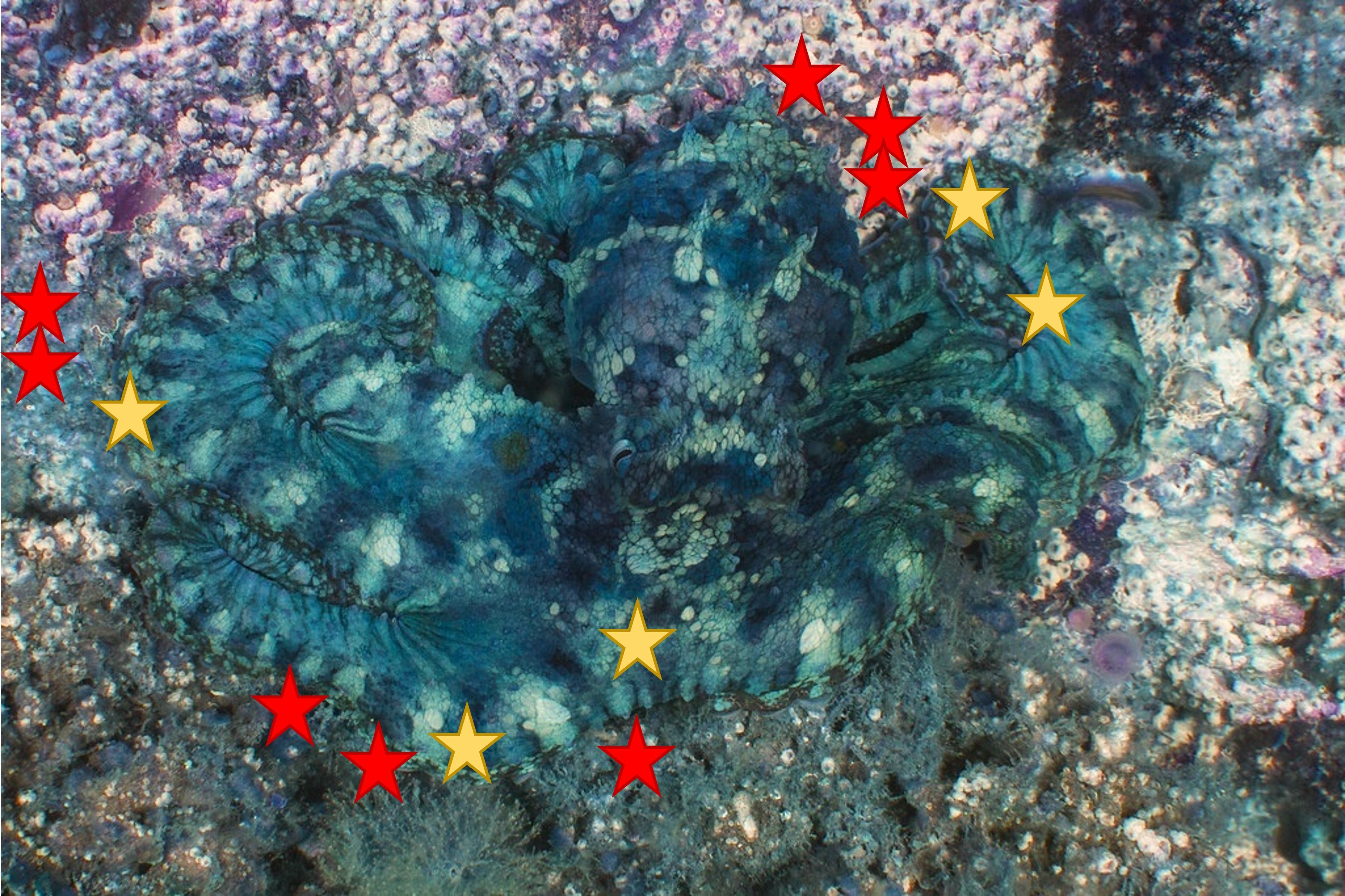}
  \label{fig:short-a}
 \end{subfigure}
 \caption{Example of prompt.{\huge \textcolor{yellow}{$\star$}} represents Positive Point and {\huge \textcolor{red}{$\star$}} represents Negative Point}
 \label{fig1}
\end{figure*}

\section{Introduction}
Camouflaged Object Detection (COD) \cite{ref17} aims to discover and figure out the objects of interest hidden in environment, whose appearances fit well into the surrounding context in terms of color and texture, and are thus hard to be perceived. Compared with general image segmentation tasks, COD is challenging in that the difference between object and environment is minor on account of the ambiguous boundary as well as the homogeneous texture and color, some examples are shown in Fig.\ref{fig1}. Thus, the general image segmentation models focused on pixel-level distinction are in general not applicable to the COD problem. This gives chance to professional COD models that are designed for perceiving specific objects, even visually indistinguishable against environment to trick human eyes. 

The existing COD models mostly rely on complex architecture design as well as subtle feature rendering to highlight unnatural transition between objects and background in terms of edge, texture, or color. For example, Sinet \cite{ref17} designs a feature processing module inspired by biological vision to detect camouflaged targets; FEDER \cite{ref19} employs wavelet decomposition to identify specific target features; \cite{05} proposes a method for detecting subtle differences between targets and backgrounds by examining features in different frequency domains of the image. In general, the existing models are very complex and not generalizable, and design of them relies on very specific professional knowledge such as wavelet analysis and frequency domain analysis. In such a context, it is hard to verify each component’s rationality, which relies mostly on the vision as well as the expertise of the designers, and thus prevents the advance of the literature. The state of the art calls for new paradigms for COD, which gives rise to big model-driven researches.

In the new era of AI, the emerging of big model brings in a new vision to rebuild the solutions for many down-stream tasks in an efficient manner, for which only minor effort is needed through promotion-based domain adaption. Segment Anything Model (SAM) \cite{ref1} is a big model with over 600 million parameters trained on huge data with over 1 billion masks from more than 11 million images, which possess very powerful segmentation capability and good generalizable ability to support broad-spectrum applications such as medical imaging \cite{12,13}, cell segmentation \cite{11}, and image editing \cite{16}.


\begin{figure}[t]
 \centering
 \includegraphics[width=4in]{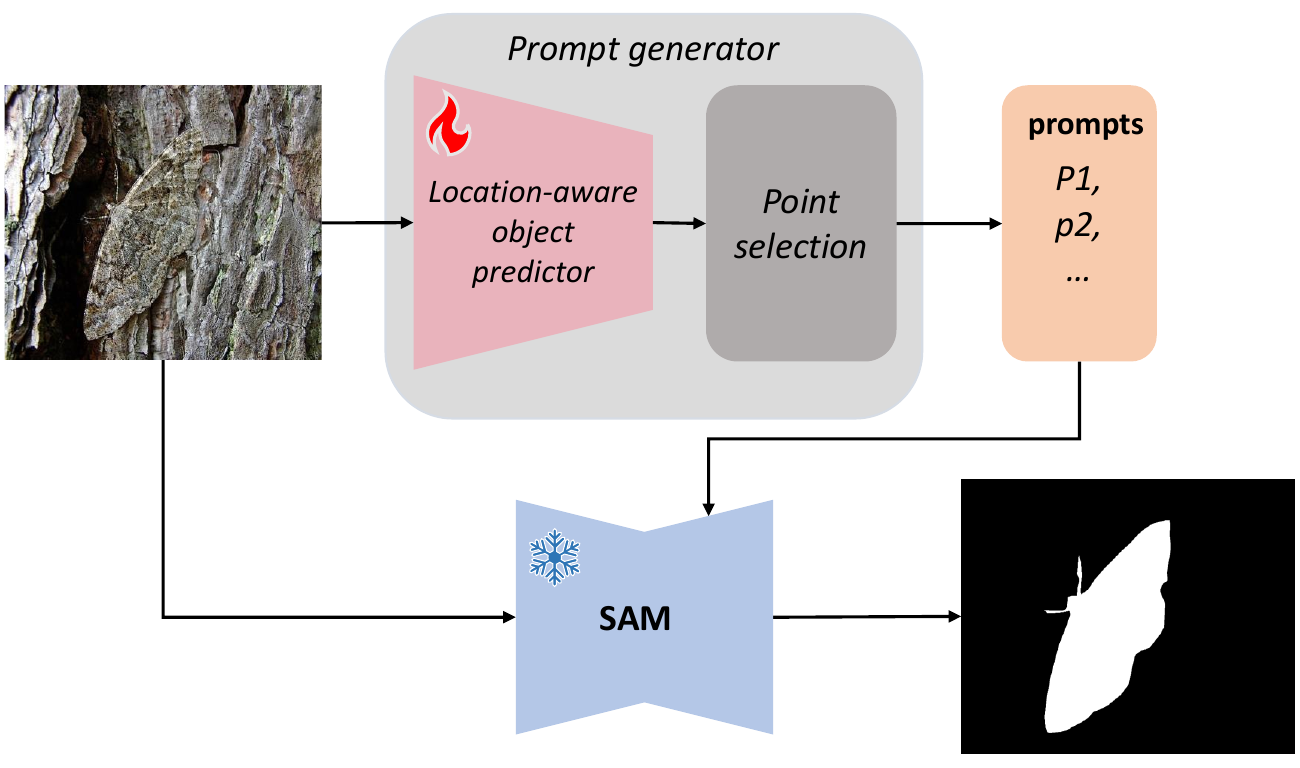}
 \caption{Overall Framework. We generate prompt points based on the input image. SAM generates a camouflaged object segmentation mask according to the image and prompt. The weights of SAM are frozen. }
 \label{fig0}
\end{figure}

Due to the challenging nature of COD, the previous studies \cite{ref3,ref5} concluded that SAM as a generic model is not applicable to COD. This study further explores such issue. Our goal is to guide SAM to segment camouflaged objects from background via prompt, where an intuition is to locate the objects roughly leveraging another network so as to promote SAM to approach the objects given some clues. We generate the promotion using a light-weight design referred to as Point Promotion Targeting Network (PPT-net) and Key Point Selection (KPS) algorithm, working in collaboration to locate the object of interest roughly. Since the promotion is just a coarse indication of the object’s position, not necessary to be precise, so the network can be made simple and light weighted while the heavy working load is assigned to SAM, taking advantage of its powerful segmentation capability in diverse scenarios. SAM can segment objects based on shape, texture, color, and semantic information, while it can accept not only image as input but also provides an interface to input promotion in the forms of bounding box, mask, textual description, and key points with positive and negative labels as guidance.

Here, we generate key point based promotion to SAM via the following scheme: At first, we generate a couple of candidate points uniformly distributed across the image. Then, we utilize the proposed PPT-net to predict the probability that an object may appear in each location by using multi-scale features obtained from the image. Then, based on the predicted probabilities, the KPS algorithm is applied to search the whole receptive field to select some representative and mutually complementary key points as positive and negative clues. Finally, such selective key point based promotions are fed to SAM to produce image segmentation. Throughout the whole pipeline, SAM is frozen such that the COD relies fully on SAM’s original segmentation power. According to the experimental results on 3 data sets under 6 metrics, the proposed solution based on PPT-net and KPS algorithm can approach the state-of-the-art (SOTA) performance in some metrics while outperforms the baselines in some other metrics to achieve new SOTA.

Our contribution is summarized as follows: 
\begin{itemize}
  \item The previous studies indicated that SAM is not applicable to COD problem but this study turns over such conclusion. In the new era highlighted by big models, this is the first investigation proposing to revisit the COD problem by leveraging big model’s strong capability with only minor effort to generate promotions through a simple model. This leads to a new methodology that enables reaching a sound solution efficiently.
  \item We propose PPT-net and KPS algorithm to obtain selective key point based promotion to SAM.
  \item The propose solution has achieved SOTA performance in some scenarios in the extensive experiments on 3 benchmarks under 6 metrics.
\end{itemize}

\section{Related Works}
\subsection{Camouflaged Object Detection (COD)}
The Camouflaged Object Detection (COD) task aims to identify and segment objects that are concealed in the environment, such as an Arctic fox in the snow or a dead leaf butterfly in fallen leaves. Camouflaged objects often have similar colors and textures in contrast to the background, and their edges are blurry and hard to be distinguished. The segmentation models for general tasks struggle to recognize and segment camouflaged objects.

\cite{ref17,ref26,01} propose COD10k, NC4K, and CAMO datasets for camouflaged object detection. These datasets include images of camouflaged objects along with corresponding segmentation masks, providing a foundation for solving COD task. Many approaches for COD tasks rely on specially designed feature processing modules and carefully crafted network architectures to detect subtle feature differences of camouflaged objects in images.

SiNeV2 \cite{ref18} simulate biological vision by progressively searching and locating camouflaged objects. BGNet and BSA-Net are focused on detecting and extracting target edge information. \cite{05} detects subtle differences between targets and backgrounds by examining features in different frequency domains. Several methods also combine different architectures and domain knowledge to render a solution: FEDER \cite{ref19} introduces wavelet decomposition to identify specific target features; MGL \cite{02} uses an additional graph neural network to detect edge information; FSPNet \cite{ref25} combines transformer architecture with GNN and employs a special feature fusion strategy; SegMaR \cite{ref20} proposes an iterative framework to address the challenging task of detecting small objects; ZoomNet \cite{ref21} scales images to obtain multi-scale features; \cite{03} introduces adversarial learning for COD tasks; OCE-Net \cite{ref23} detects camouflaged objects by using the uncertainty of the data itself.

These existing solutions incorporate specialized modules and network architectures tailored to the characteristics of COD tasks, making it challenging to transfer these methods to other segmentation tasks. Additionally, some works introduce domain knowledge, which can prevent easy understanding and subsequent research efforts.
\subsection{Segment Anything Model (SAM)}
SAM (Segment Anything Model) \cite{ref1} is a segmentation model with powerful segmentation capability and sound generalization. It can accept additional inputs in a specific form as prompt, typically in the form of coordinate points and bounding boxes, to guide the model to segment some specific target objects. By scoring the quality of segmentation outcome to incorporate high-quality results into a training corpus with more than one-billion segmentation masks, the segmentation and generalization capability of SAM becomes promising.

Due to SAM's powerful performance as well as its interface to enable further improvement through user-defined prompt, many related works have emerged in the literature. SAMAug \cite{ref2} iteratively generates point-based prompts according to segmentation results to improve SAM's segmentation performance, where the prompts are generated through an additional saliency detection network to impose restriction on the ranges of the coordinates in iterating the process, in order to compensate for the possibly unexpected holes or structural deficiencies on SAM generated segmentation masks.

Some works are focused on applying SAM in specific tasks, such as medical imaging \cite{12,13}, cell segmentation \cite{11}, image editing \cite{16}, crack detection \cite{17}, crater detection \cite{18}, and remote sensing imagery \cite{110}. However, SAM does not consistently perform well on all tasks. \cite{ref3,ref5} indicates that SAM is not effective in completing COD tasks, due to the diverse scenarios as well as the challenging nature of COD tasks. Our work overcomes this limitation and significantly improves SAM's segmentation performance on COD tasks.

The concept of prompt originating from the natural language processing (NLP) field \cite{21} has been extended to the computer vision (CV) field, commonly referred to as visual prompt (VP) in the visual domain. Prompt-based methodology provides specific additional inputs for a model without altering the existing model’s parameters or architecture such that the model parameters are frozen to preserve its original performance, and meanwhile, it can enable promoting a model to accomplish other specific tasks with improved performance. The forms of prompts can vary, including text descriptions \cite{22,23}, special noises, and structured data. 

Learning how to generate high-quality prompts to guide models toward expected goals is also a major concern. As for using specific parameters as prompts, \cite{ref10} and \cite{ref14} learn to generate specific noises as prompts along with images to enhance the model's performance when transferred to down-stream classification tasks. Concerning Transformer architectures, \cite{ref11} proposes using prompts as additional tokens as the encoder's input to enhance feature distinctiveness, while \cite{ref12} applies additional embeddings within the encoder as prompts. \cite{ref9} focuses on generating prompts to improve the segmentation performance of models in low-level tasks. Some works explore the generation of prompts for continual learning to continuously improve model performance via post-deployment. For Test-Time Adaptation, \cite{ref15} proposes a method of prompt learning that supports continuous learning to enhance the model's robustness. Additionally, some methods use user-generated prompts for interactive purposes. For example, \cite{ref8} introduces an iteratively interactive segmentation approach, using user-annotated coordinate points as additional inputs to segment the special object of interest.

Although SAM can generate specific segmentation results based on manually annotated prompts (such as text descriptions, coordinate points, or bounding boxes), this manual labeling approach is inefficient. Our work addresses this issue by implementing automatic generation of high-quality prompts while preserving SAM's original performance to serve COD task.

\section{Method}
Our goal is to produce triples of points as the promotion to SAM, in each of which one point provides a positive clue indicating the possible existence of a camouflaged object while the other two coupled points indicate the negative cases in a contrastive way. Then, SAM will be promoted to figure out the full profiles of the camouflaged objects under the guidance of the promotion triple points, during which SAM is frozen, and thus selection of key points for promoting SAM plays a very critical role for COD. The overview of the principle is illustrated in Fig.\ref{fig0}. 

\begin{figure*}[t]
 \centering
 \label{fig-1}
  \includegraphics[scale=0.5]{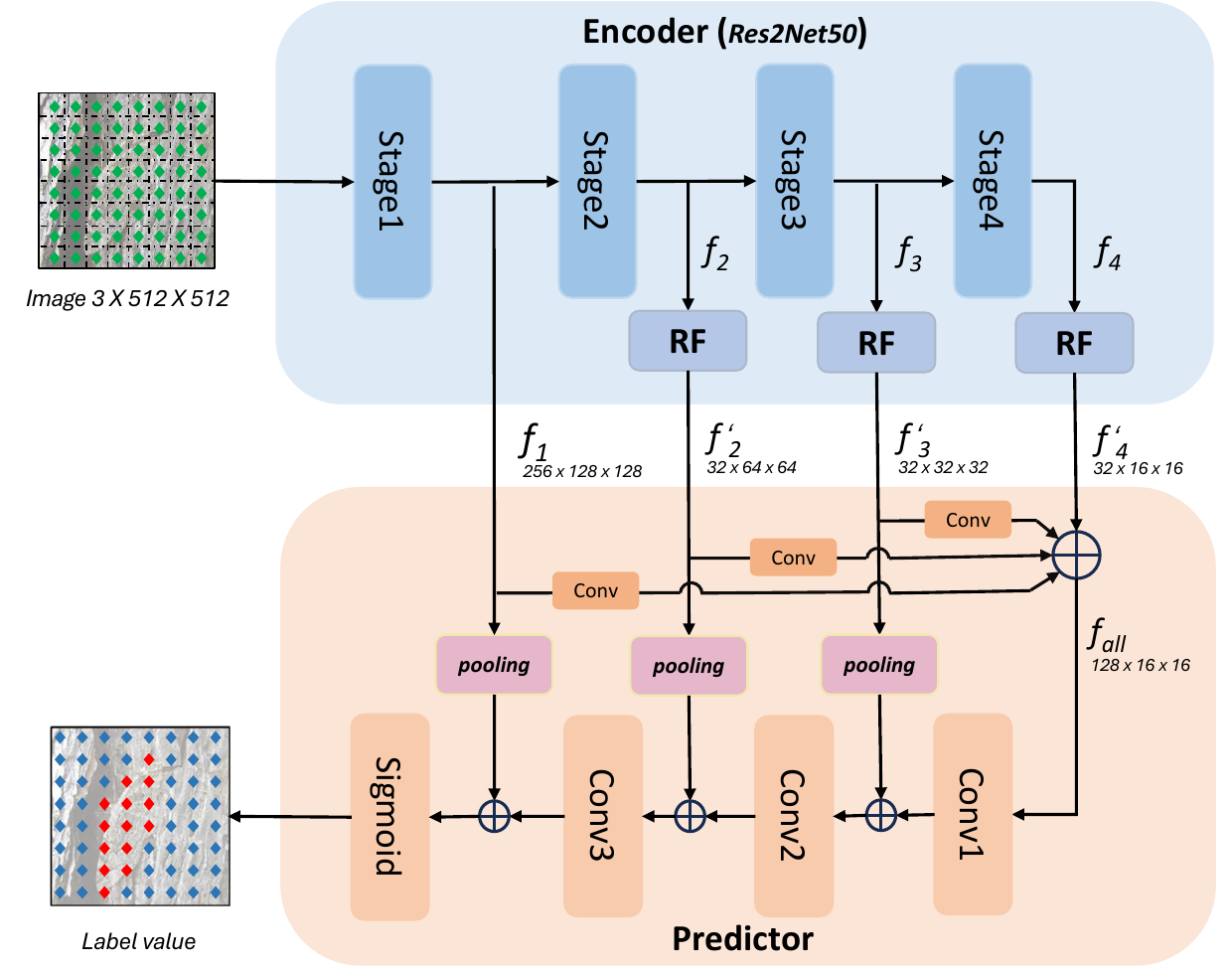}
 \caption{Overall architecture of PPT-net. It consists of two key components: The encoder to extract multi-scale features, and the predictor to output the probability of the presence of camouflaged targets at each given point.}
 \label{fig2}
\end{figure*}
\subsection{Visual Prompt}

Note that SAM can accept prompt in the form of point, box, mask, and textural description but in this case, point based promotion is a wise choice, because this degrades the challenge of direct COD into figuring out only a coarse profile of camouflaged objects, not necessary to be complete, nor precise. Considering the scenario that the appearances of unknown number of camouflaged objects fit closely into the background with ambiguous boundary and an object may be partially blocked by other items to fall into separated parts, in such a case, key points should be more robust than the other form of promotions. SAM can accept two types of point based promotions, positive or negative, which indicate possible existence or inexistence of targets, respectively. In this study, we apply both positive promotion (PP) and negative promotion (NP) in the form of a triple (1 PP against 2 NP) to indicate objects and background in a contrastive way to promote SAM.

Since predicting the exact positions of promotion points is difficult, inspired by YOLO \cite{ref27}, we do not predict the coordinates of points directly, but instead, we predict the probability of each given point of interest as positive or negative. That is, we first generate some unlabeled points and then predict the probability whether a point could be positive/negative. Finally, we select some key points from such candidates as PP or NP according to the predicted probability in association with each candidate point.

The overall solution is composed of 3 pipelined modules: (1) Mark the input image with M$\times$M grid partition-minded locality so as to render the center of each box as an unlabeled candidate point, by which each candidate point is associated with a box-represented locality of the corresponding image part, and in total, N = M$\times$M candidate points are obtained. (2) Predict the probability that each candidate point associates with camouflaged objects using the proposed Promotion Point Targeting Network (PPT-net). (3) Apply the so-called Key Point Selection (KPS) algorithm to select PP-NP-NP triples as the promotion to SAM. The implementation of PPT-net and KPS algorithm are detailed in the following.
\subsection{Promotion Point Targeting Network (PPT-net)}
The PPT-Net is composed of two components: Encoder and predictor, as shown in Fig.\ref{fig2}. The encoder leverages Res2Net as the backbone to extract multi-scale features from the input image $X$, where the 4 scales of features denoted as $f_1, f_2, f_3,$ and $f_4$ represent from low-level to high-level abstractive features, respectively. Then, the multi-scale features are concatenated after passing through Receptive Field (RF) Block \cite{ref18} and convolution (Conv) block to augment their discriminant power towards COD as in SinetV2 \cite{ref18}. The aggregated feature fed to the predictor is:
\begin{align}
\begin{split}
f_{all} =  Conv(f_1)\oplus Conv(RF(f_2))
      \oplus Conv(RF(f_3))\oplus RF(f_4).
\end{split}
\end{align}

\noindent where $ \oplus $ means concatenation, and $f_1$ is an exception not undergoing RF block so as to preserve the fundamental low-level feature in terms of texture, color, and edge. The predictor is composed of 3-layer stacked convolutions with sigmoid-activated output layer. During the inference of the predictor, we apply both the global feature ($f_{All}$) and the multi-scale local features arising from the maximum pooling performed on the M$\times$M grid partition of the feature maps, where the grid is scaled to adapt to the size of the corresponding feature map. Then, $f_{All}$ is applied as input to the predictor while the multi-scale features with regard to each block in association with a candidate point are applied as outside controls, injected into the intermediate pipelines in turn. Thus, the prediction is driven by both the global feature ($f_{All}$) and the local features 
\begin{equation}
\begin{split}
     \{ MP_i(f_1),\ MP_i(RF(f_2)),\ 
     MP_i(RF(f_3))|i=1,2,…,N\} 
\end{split}
\end{equation}
where $MP_i$ denotes maximum pooling on the block centering on the corresponding unlabeled candidate point $C_i$. Such a procedure coincides with the intuition that camouflaged objects refer to not only local measure but also its context, say, global feature. 

In training, we use the binary ground-truth mask of each image to train PPT-net, where pixel value 1 of the mask indicates camouflaged object and 0 the background. The pixel values of each block of the ground-truth mask is averaged to score the presence of camouflaged objects in terms of probability with regard to the centroid of this block, and PPT-net is taught to predict such probability of each candidate point through the MSE loss defined as follows: 
\begin{equation}
      Loss= MSE(PPT\text{-}net(X),AvgPooling(Mask))
\end{equation}

\noindent where $X$ denotes the input image, and AvePooling the average pooling over the window centering on a candidate point $C_i$, in correspondence with the grid partition rendered windows and the center within each window. Then, the values predicted from PPT-net can be associated with the corresponding candidate points in the form of\\
\begin{equation}
      (C_i,p_i) : i=1,2,...,N.
\end{equation}

\noindent where $N$ represent the number of pre-generated points, also the number of windows following grid partition, and\\
\begin{equation}
      (p_1,p_2,...,p_N) = PPT\text{-} net(X)
\end{equation}

\noindent the probability to measure how far point $C_i$ is possibly within camouflaged objects, $i$=1, 2, ..., $N$.

\subsection{Key Point Selection (KPS) Algorithm}

The predicted scores of PPT-net indicate the probability of the presence of camouflage objects at each position of interest. We consider high prediction score to indicate the presence of camouflage objects at a point, while low prediction score represent the background. According to the KPS algorithm, we select points with high confidence and representativeness as prompts. We select the first PP as the point with the highest prediction score, and then add in turn the candidate point with the maximum sum of the distances to the existing PPs as a new member to the set of PPs so as to represent the overall outline of camouflage objects. Finally, we select the two corresponding points with the minimum distances to each PP point as the two associated NPs to provide contrastive information indicating roughly the edge of the objects.

According to the predicted score, we sort the candidate points into 4 levels:

\begin{equation}
\begin{split}
    L_1&= \{ C_i \mid p_i>0.9, i \in [1,N] \}; \\
    L_2&= \{ C_i \mid 0.75 \leq p_i \leq 0.9, i \in [1,N]\};\\
    L_3&= \{ C_i \mid 0.5 \leq p_i<0.75, i \in [1,N] \}; \\
    L_4&= \{ C_i \mid p_i<0.5, i \in [1,N] \}; 
\end{split}
\end{equation}

\noindent $L_1$ contains points with very high confidence to indicate camouflage objects and can thus act as reliable positive prompts for SAM. $L_2$ contains points with relatively high confidence, ensuring adequate representation of object outlines while maintaining sound prompts. $L_3$ contains a large number of points with moderate confidence scores, making it ambiguous as either positive or negative prompts. $L_4$ contains points with low prediction scores, indicating that they should be considered as background. Since the KPS algorithm pairs each PP with 2 NP points and requires NP points near the edges, the two NPs in association with a PP should be selected from $L_4$ with the minimum distances to the PP, that is, the closest background points in collaboration with the foreground point to indicate roughly the edge in a contrastive way.

At first, we select PP from $L_1$ and $L_2$, and then NP from $L_4$. In searching PP, we select the points locating as far as possible from each other to cover the whole receptive filed extensively, and $L_1$ has the priority to be searched. In case $L_1$ has less than $K$ PPs, a user predefined number, then, we search for the remaining PPs in $L_2$. If $L_1$ and $L_2$ are both empty, we select the point with the highest predicted score from $L_1\cup L_2\cup L_3\cup L_4$ as PP. 

After obtaining the PPs, we search $L_4$ to find the 2 nearest neighbors of each PP as the two coupled NPs to it. The algorithm is summarized in Algorithm.\ref{alg:algorithm}. In the pseudo codes, d($C_i,P_j$) means distance between two points $C_i$ and $P_j$ while d($C_i$,$P$) the distance from point $C_i$ to the collection of the points in set $P$. 

\begin{algorithm}[h]
  \caption{Key Point Selection (KPS)}
  \label{alg:algorithm}
  \textbf{Input:} $L_1, L_2, L_3, L_4, K$ \\
  \textbf{Output:} $\text{PNN} = \{(P_1, N_{i1}, N_{i2}) \mid 1 \leq i \leq K\}$
  
  \begin{algorithmic}[1] 
    \STATE $P \gets \emptyset$; $\text{PNN} \gets \emptyset$; $L \gets L_1 \cup L_2 \cup L_3 \cup L_4$;
    \STATE /* Select positive points */
    \WHILE{$|L_1| > 0$ \textbf{and} $|P| < K$}
      \STATE $k \gets \arg\max \{ d(C_i, P) \mid C_i \in L_1 \}$;
      \STATE $P \gets P \cup \{C_k\}$, $L_1 \gets L_1 \setminus \{C_k\}$;
    \ENDWHILE
    \WHILE{$|L_2| > 0$ \textbf{and} $|P| < K$}
      \STATE $k \gets \arg\max \{ d(C_i, P) \mid C_i \in L_2 \}$;
      \STATE $P \gets P \cup \{C_k\}$, $L_2 \gets L_2 \setminus \{C_k\}$;
    \ENDWHILE
    \IF{$P$ is empty}
      \STATE $k \gets \arg\max_{C_i \in L} (p_i = \text{PPT-net}(C_i))$;
      \STATE $P \gets P \cup \{C_k\}$, $L \gets L \setminus \{C_k\}$;
    \ENDIF
    \STATE /* Select negative points */
    \FOR{$j = 1$ \textbf{to} $|P|$ \textbf{and} $|L_4| > 0$}
      \STATE $k_1 \gets \arg\min \{ d(C_i, P_j) \mid C_i \in L_4 \}$;
      \STATE $k_2 \gets \arg\min \{ d(C_i, P_j) \mid C_i \in L_4 \text{ and } i \neq k_1 \}$;
      \STATE $\text{PNN} \gets \text{PNN} \cup \{(P_j, C_{k1}, C_{k2})\}$, $L_4 \gets L_4 \setminus \{C_{k1}, C_{k2}\}$;
    \ENDFOR
    \STATE \textbf{return} $\text{PNN}$
  \end{algorithmic}
\end{algorithm}

\subsection{Camouflaged Object Detection}
Through PPT-net and KPS algorithm, we can obtain a couple of triple points as positive and negative promotions to SAM. Here, we feed the input image and the point promotions to SAM in a way like SAMaug \cite{ref2}, that is, once we obtain the initial image segmentation from SAM via the key point based promotion, we apply the outcome as mask promotion and reboot SAM with both mask and key point based promotions to produce a new segmentation as final output.
\section{Experiments}
\subsection{Experimental Setup}

\linespread{1.5}
\begin{table*}[t]
 \centering
 \resizebox{\textwidth}{!}{
 \begin{tabular}{l|cccccc|cccccc|cccccc}
\toprule
\multirow{2}{*}{Methods} & \multicolumn{6}{c|}{COD10K (2,026)} & \multicolumn{6}{c|}{NC4K (4,121)} & \multicolumn{6}{c}{CAMO (250)} \\ 
\cline{2-19}  
& $S_m\uparrow$ & $F^w_\beta\uparrow$ & $F^m_\beta\uparrow$ & $E^m_\phi\uparrow$ & $E^x_\phi\uparrow$ & $M\downarrow$ 
& $S_m\uparrow$ & $F^w_\beta\uparrow$ & $F^m_\beta\uparrow$ & $E^m_\phi\uparrow$ & $E^x_\phi\uparrow$ & $M\downarrow$ 
& $S_m\uparrow$ & $F^w_\beta\uparrow$ & $F^m_\beta\uparrow$ & $E^m_\phi\uparrow$ & $E^x_\phi\uparrow$ & $M\downarrow$ \\ 
\midrule
FEDER & $0.820$ & $0.714$ & $0.748$ & $0.897$ & $0.905$ & $0.032$ 
& $0.827$ & $0.764$ & $0.803$ & $0.889$ & $0.900$ & $0.052$ 
& $0.799$ & $0.732$ & $0.774$ & $0.861$ & $0.869$ & $0.073$ \\
SiNetV2 & $0.815$ & $0.679$ & $0.717$ & $0.886$ & $0.906$ & $0.037$ 
& $0.847$ & $0.769$ & $0.803$ & $0.901$ & $0.913$ & $0.048$ 
& $\mathbf{0.820}$ & $0.742$ & $0.781$ & $\mathbf{0.881}$ & $\mathbf{0.895}$ & $0.070$ \\
ZoomNet & $\mathbf{0.838}$ & $0.729$ & $0.766$ & $0.888$ & $\mathbf{0.911}$ & $\mathbf{0.029}$ 
& $\mathbf{0.853}$ & $0.784$ & $0.818$ & $0.896$ & $0.912$ & $\mathbf{0.043}$ 
& $\mathbf{0.820}$ & $\mathbf{0.752}$ & $0.794$ & $0.877$ & $0.892$ & $\mathbf{0.066}$ \\
BGNet & $0.829$ & $0.710$ & $0.753$ & $\mathbf{0.900}$ & $0.910$ & $0.033$ 
& $0.851$ & $0.780$ & $0.819$ & $\mathbf{0.907}$ & $\mathbf{0.916}$ & $0.045$ 
& $0.812$ & $0.743$ & $0.789$ & $0.871$ & $0.882$ & $0.074$ \\
OCE-Net & $0.825$ & $0.704$ & $0.738$ & $0.892$ & $0.905$ & $0.033$ 
& $0.851$ & $0.779$ & $0.811$ & $0.899$ & $0.911$ & $0.046$ 
& $0.802$ & $0.722$ & $0.764$ & $0.850$ & $0.865$ & $0.081$ \\
BSA-Net & $0.817$ & $0.699$ & $0.737$ & $0.889$ & $0.902$ & $0.034$ 
& $0.841$ & $0.771$ & $0.806$ & $0.895$ & $0.907$ & $0.048$ 
& $0.794$ & $0.717$ & $0.760$ & $0.850$ & $0.866$ & $0.079$ \\
SiNet & $0.775$ & $0.629$ & $0.680$ & $0.863$ & $0.875$ & $0.043$ 
& $0.808$ & $0.721$ & $0.769$ & $0.871$ & $0.883$ & $0.058$ 
& $0.747$ & $0.645$ & $0.703$ & $0.804$ & $0.830$ & $0.092$ \\
SegMaR & $0.754$ & $0.568$ & $0.626$ & $0.811$ & $0.860$ & $0.060$ 
& $0.766$ & $0.633$ & $0.692$ & $0.794$ & $0.881$ & $0.089$ 
& $0.727$ & $0.590$ & $0.651$ & $0.755$ & $0.854$ & $0.116$ \\
\midrule
ours & $0.826$ & $\mathbf{0.750}$ & $\mathbf{0.788}$ & $0.887$ & $0.890$ & $0.032$ 
& $0.836$ & $\mathbf{0.789}$ & $\mathbf{0.825}$ & $0.885$ & $0.888$ & $0.049$ 
& $0.790$ & $0.742$ & $\mathbf{0.790}$ & $0.845$ & $0.847$ & $0.079$ \\ 
\bottomrule
 \end{tabular}
 }
\caption{Performance comparison with baselines. }
\label{table-2}
\end{table*}

\begin{table}[t]
    \centering
    \resizebox{0.97\textwidth}{!}{
        \begin{tabular}{lcccccc|cccccc}
            \toprule
            \multirow{2}{*}{Methods} & \multicolumn{6}{c}{NC4K} & \multicolumn{6}{c}{COD10K}\\ 
            \cline{2-13}   
            & $S_m\uparrow$ & $F^w_\beta\uparrow$ & $F^m_\beta\uparrow$ & $E^m_\phi\uparrow$ & $E^x_\phi\uparrow$ & $M\downarrow$ 
            & $S_m\uparrow$ & $F^w_\beta\uparrow$ & $F^m_\beta\uparrow$ & $E^m_\phi\uparrow$ & $E^x_\phi\uparrow$ & $M\downarrow$\\ 
            \midrule
            Baseline (SiNet) & 0.808 & 0.721 & 0.769 & 0.871 & 0.883  & 0.058 
            & 0.775 & 0.629 & 0.680 & 0.863 & 0.875  & 0.043\\
            SAM & 0.406 & 0.034 & 0.040 & 0.338 & 0.339 & 0.186 
            & 0.445 & 0.015 & 0.018 & 0.315 & 0.315 & 0.111\\ 
            SAM+prompt & \textbf{0.836} & \textbf{0.789} & \textbf{0.825} & \textbf{0.885} & \textbf{0.888} & \textbf{0.049} 
            & \textbf{0.826}  & \textbf{0.750} & \textbf{0.788} & \textbf{0.887} & \textbf{0.890} & \textbf{0.032}\\
            \bottomrule
        \end{tabular}
    }
    \caption{Performance improvement arising from the proposed prompt}
    \label{table-1}
\end{table}

we evaluate our method on three widely used public datasets: COD10K \cite{ref17}, NC4K \cite{ref26}, and CAMO \cite{ref28}. COD10K is currently the largest COD dataset, comprising 5,066 camouflage images, 3,000 background images, and 1,934 non-camouflage images. NC4K is the latest large COD test dataset, containing 4,121 images. CAMO consists of 1,250 camouflage images and 1,250 non-camouflage images. Following the previous works \cite{ref18,ref22,ref19}, we train our method only on COD10K data set and resize the resolution of each image to 3$\times$512$\times$512.

For the baselines, we use directly the open-source codes provided by the authors along with their parameter settings. SAM is frozen throughout the whole procedure of training without any fine tune, and in our implementation, the weights as well as codes are adapted from \cite{ref1} directly. For the encoder of PPT-net, its weights are adapted from SinetV2 \cite{ref18} directly.

We use Adam optimizer with an initial learning rate of 0.001, and apply dynamic learning rate decay. The batch size is 32 and the training ends after 300 epochs. All the experiments were conducted on a single NVIDIA RTX 3090.

As for performance evaluation, we use six commonly used metrics: S-measure (S$_m$) \cite{ref31}, weighted F-measure (F$^w_\beta$) \cite{ref32}, mean F-measure (F$^m_\beta$) \cite{ref33}, mean E-measure (E$^m_\phi$) \cite{ref34}, max E-measure (E$^x_\phi$), and mean absolute error (M). Smaller values are better for M, while larger values are preferable for the other metrics.

\begin{table}[t]
    \centering
    \resizebox{0.67\textwidth}{!}{
        \begin{tabular}{cccccccc}
            \toprule
            & \#Candidates & $S_m\uparrow$ & $F^w_\beta\uparrow$ & $F^m_\beta\uparrow$ & $E^m_\phi\uparrow$ & $E^x_\phi\uparrow$ & $M\downarrow$\\ 
            \midrule
            & $32\times32$ & \textbf{0.834} & \textbf{0.763} & \textbf{0.799} & \textbf{0.894} & \textbf{0.897} & \textbf{0.031}\\ 
            & $16\times16$ & 0.826  & 0.750 & 0.788 & 0.887 & 0.890 & 0.032\\
            \bottomrule
        \end{tabular}
    }
    \caption{Performance variation against the number of unlabeled candidate points.}
    \label{table-3}
\end{table}
\subsection{Comparison with Baselines}
As shown in Table \ref{table-2}, our method can achieve competitive and sometimes superior performance against the SOTA baselines, including FEDER \cite{ref19}, ZoomNet \cite{ref21}, BGNet \cite{ref22}, OCE-Net \cite{ref23}, BSA-Net \cite{ref24}, SiNetV2 \cite{ref18}, SiNet \cite{ref17}, and SegMaR \cite{ref20}. Notably, our approach establishes a new state-of-the-art in terms of F$^m_\beta$ on all the 3 datasets, while performs best in terms of F$^w_\beta$ on COD10K and NC4K. 

The high F$^w_\beta$ and F$^m_\beta$ scores indicate that our method performs well when considering both precision and recall, excellent in capturing the overall shape of camouflage objects.

In terms of E$^m_\phi$ and E$^x_\phi$ that are focused on measuring the exactness of the edge, our method does not outperform the existing methods due to the finite number of prompt points that are insufficient to cover all the edges of camouflage objects.

In the S$_m$ and M metrics, the performance of our method is close to SOTA, where the slight differences possibly arise from such a case that some camouflage objects fall into multiple isolated parts due to obstacles, out of the capability of SAM. 

Nevertheless, the experiments demonstrate that a well-designed prompting mechanism allows SAM to achieve promising results on untrained tasks.

\subsection{Ablation Study}
We check the impact of each key factor on the full solution, including with/without promotion, grid partition, number of promotion points, and working pipeline to leverage SAM.

\begin{table}[t]
    \centering
    \resizebox{0.67\textwidth}{!}{
        \begin{tabular}{ccccccccc}
            \toprule  
            & \#PP & \#NP & $S_m\uparrow$ & $F^w_\beta\uparrow$ & $F^m_\beta\uparrow$ & $E^m_\phi\uparrow$ & $E^x_\phi\uparrow$ & $M\downarrow$\\ 
            \midrule 
            & 3 & - & 0.787 & 0.694 & 0.745 & 0.838 & 0.841 & 0.039\\ 
            & 4 & - & 0.788 & 0.696 & 0.746 & 0.840 & 0.842 & 0.039\\
            & 5 & - & 0.823 & 0.747 & 0.786 & 0.884 & 0.887 & 0.033\\
            & 5 & 10 & \textbf{0.826} & \textbf{0.750} & \textbf{0.788} & \textbf{0.887} & \textbf{0.890} & \textbf{0.032}\\
            \bottomrule
        \end{tabular}
    }
    \caption{Impact of the Number of Promotion Points}
    \label{table-4}
\end{table}

\begin{table}[t]
    \centering
    \resizebox{0.67\textwidth}{!}{
        \begin{tabular}{cccccccc}
            \toprule  
            & \#Iterations & $S_m\uparrow$ & $F^w_\beta\uparrow$ & $F^m_\beta\uparrow$ & $E^m_\phi\uparrow$ & $E^x_\phi\uparrow$ & $M\downarrow$\\ 
            \midrule 
            & 1 & 0.821 & 0.746 & 0.783 & \textbf{0.888} & \textbf{0.890} & 0.034\\
            & 2 & \textbf{0.826} & \textbf{0.750} & \textbf{0.788} & 0.887 & \textbf{0.890} & \textbf{0.032}\\
            \bottomrule
        \end{tabular}
    }
    \caption{Performance Related to the Working Pipeline Utilizing SAM}
    \label{table-5}
\end{table}

\subsubsection{With/without Promotion}
We compare the performance of SAM with/without prompt, and that of the baseline SiNet \cite{ref17}. As shown in Table.\ref{table-1}, without prompt, SAM performs quite poor in comparison with SiNet. In contrast, the promotion to SAM improves its COD performance to a remarkably high level that is close to or better than SiNet, a professional model for COD only.
\subsubsection{Grid Partition}
As shown in Table.\ref{table-3}, when the number of unlabeled candidate points is improved from 16$\times$16 to 32$\times$32, the performance on the COD10K dataset improves a little as fine-grained candidate points has a higher chance to overlap with camouflaged objects.

\subsubsection{Number of Promotion Points}
We compared different configurations of the number of PP against that of NP: 3PPs-0NP, 4PPs-0NP, 5PPs-0NP, and 5PPs-10NPs. As shown in Table.\ref{table-4}, increasing either the number of PP or that of NP can improve the performance. This is because more positive promotion points allows a finer grain to cover the receptive filed, which leads to a higher change to approach camouflaged objects, while the contrastive reference provided by negative promotions do also contribution to the promoting.

\subsubsection{Working Pipeline to Leverage SAM}
As shown in Table.\ref{table-5}, we compare the case running SAM once with only point-based promotion to that running SAM twice with both the point-based promotion and the feedback resulting from the first time of running SAM as mask promotion. The experimental results show that the second case performs better in an overall sence. This means that SAM can repair its original decision on the COD problem if running recursively with the aid of point-based promotion.
\section{Conclusion}
This study demonstrates a new paradigm for Camouflaged Object Detection by leveraging big model, namely SAM, with selective key point based promotion, for which we propose the Promotion Point Targeting Network (PPT-net) along with the Key Point Select (KPS) Algorithm. The experiments on 3 benchmarks show that the proposed solution is promising and achieved the state-of-the-art performance in some scenarios.

\clearpage  

%
%
\bibliographystyle{splncs04}
\bibliography{main}
\end{document}